\documentclass[10pt,twocolumn,letterpaper]{article}

\usepackage[final]{cvpr}      

\usepackage{tcolorbox}
\tcbuselibrary{listings}
\tcbuselibrary{listingsutf8}
\tcbuselibrary{raster}
\usepackage{graphicx}
\usepackage{xcolor}
\usepackage{hyperref}
\tcbuselibrary{listings,skins,breakable,raster}
\usepackage{listings}
\usepackage{upquote}

\usepackage{soul}
\definecolor{cvprblue}{rgb}{0.21,0.49,0.74}
\definecolor{hotpink}{RGB}{255, 20, 147}
\usepackage[ruled,vlined]{algorithm2e}

\def\paperID{390} 
\def\confName{3DV\xspace}
\def\confYear{2026\xspace}

\title{Seeing Through Clutter\\ Structured 3D Scene Reconstruction via Iterative Object Removal}

\author{
Rio Aguina-Kang$^{1}$
\and
Kevin James Blackburn-Matzen$^{2}$ \and
Thibault Groueix$^{2}$ \and
Vladimir Kim$^{2}$ \and
Matheus Gadelha$^{2}$ \\
\\
$^{1}$University of California, San Diego \\
$^{2}$Adobe Research \\
}

\begin{document}
\twocolumn[{%
\renewcommand\twocolumn[1][]{#1}%
\maketitle
        \vspace{-30pt}
\begin{center}
    \centering
    \captionsetup{type=figure}
        \includegraphics[width=\textwidth]{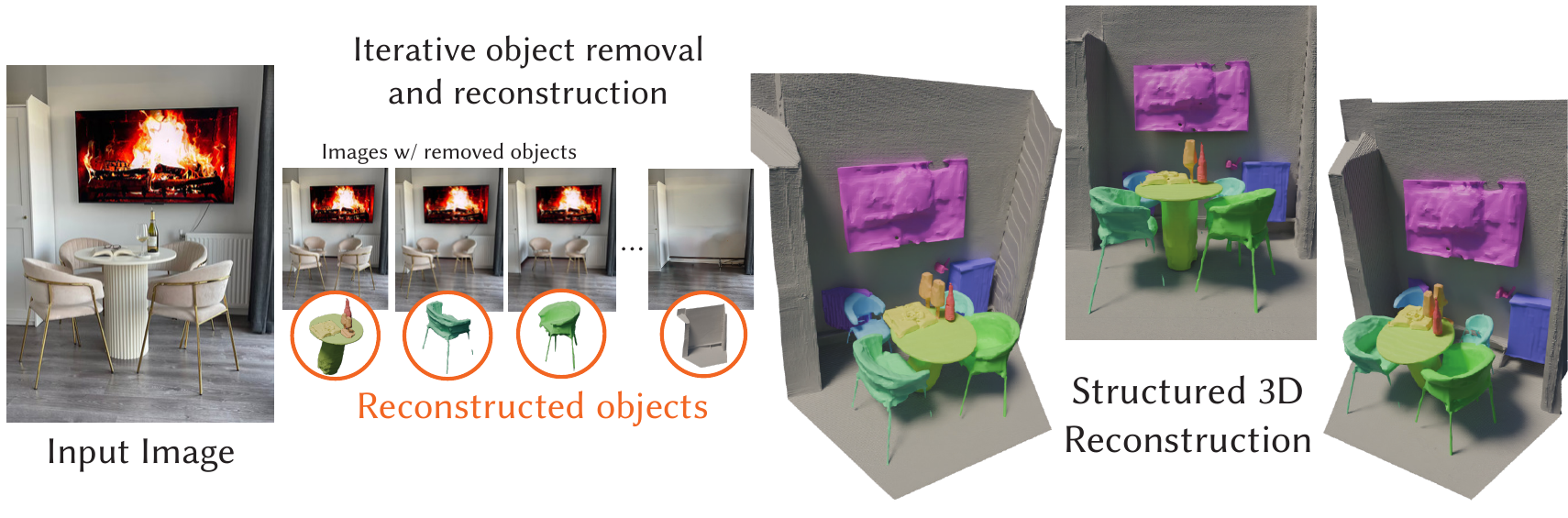}
        \vspace{-20pt}
        \caption{
            \textit{SeeingThroughClutter} is a training-free method
            that combines vision-language models (VLMs), object removal,
            monocular depth estimation, and image-to-3D reconstruction to
            automatically decompose a single photograph into a structured 3D scene.
            Given an input image, a VLM repeatedly identifies the most prominent
            foreground object, reconstructs and poses it in 3D, removes it from
            the image, and then iterates this process until no objects remain.
        }
    \label{fig:teaser}
\end{center}%
}]

\begin{abstract}
We present SeeingThroughClutter, a method for reconstructing structured 3D representations from single images by segmenting and modeling objects individually. Prior approaches rely on intermediate tasks such as semantic segmentation and depth estimation, which often underperform in complex scenes, particularly in the presence of occlusion and clutter.
We address this by introducing an iterative object removal and reconstruction pipeline that decomposes complex scenes into a sequence of simpler sub-tasks.
Using VLMs as orchestrators, foreground objects are removed one at a time via detection, segmentation, object removal, and 3D fitting. We show that removing objects allows for cleaner segmentations of subsequent objects, even in highly occluded scenes. Our method requires no task-specific training and benefits directly from ongoing advances in foundation models. We demonstrate state-of-the-art robustness on 3D-Front and ADE20K datasets. Project Page: {\color{hotpink}\href{https://rioak.github.io/seeingthroughclutter/}{https://rioak.github.io/seeingthroughclutter/}}

\end{abstract}




\section{Introduction}

\begin{figure}[t]
      \centering
      \includegraphics[width=\linewidth]{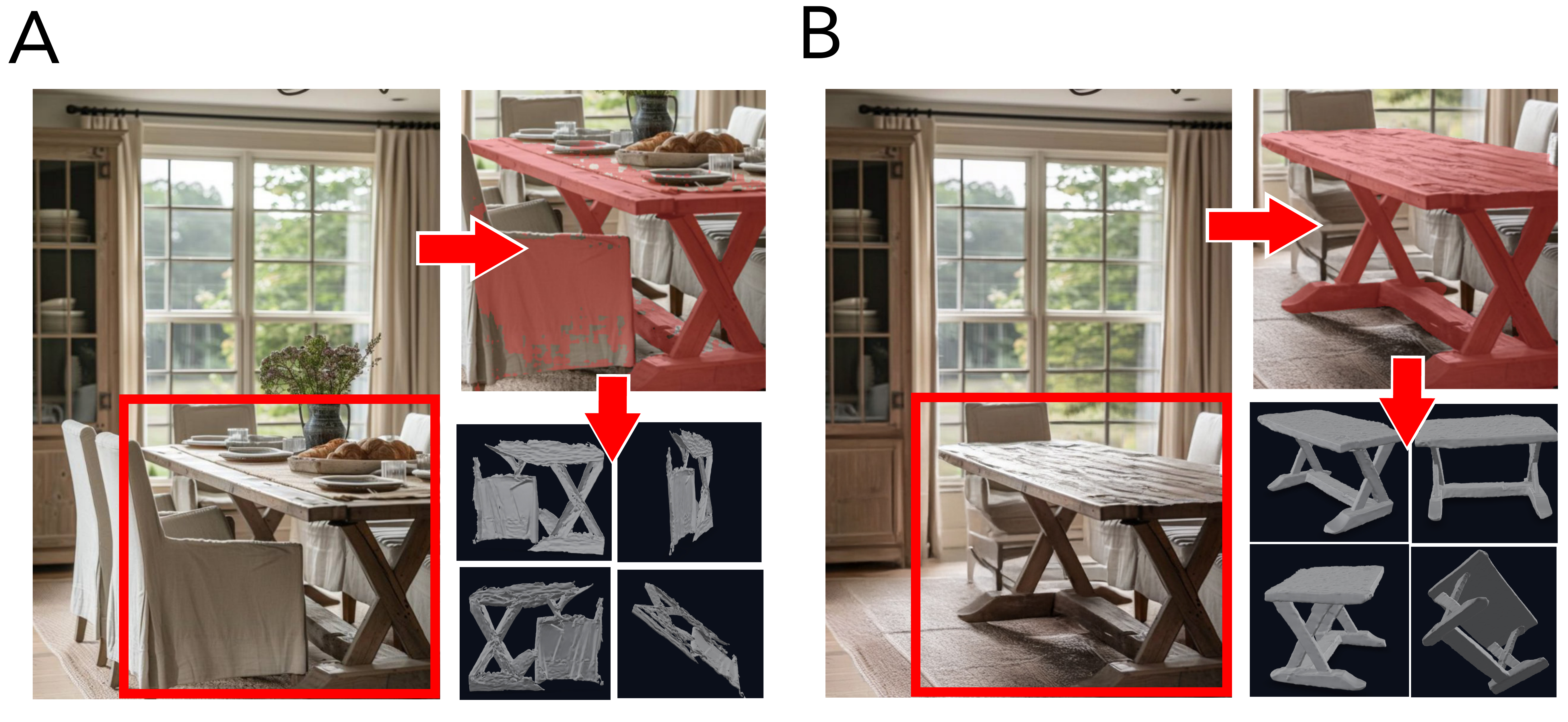}
      \vspace{-.5cm}
    \caption{Single-view scene reconstruction is often challenging due to the complexity and clutter of real-world environments. Consider a scenario where the goal is to reconstruct a table within a scene. In Figure~\ref{fig:declutter_example}A, we localize and segment the table using a bounding box; however, the resulting mask is noisy and unclear, largely because various items on the tabletop—and other objects like chairs—fall inside the box. In contrast, Figure~\ref{fig:declutter_example}B shows the same scene after these extraneous objects have been removed from the visible region, yielding a much cleaner and more accurate segmentation and reconstruction.}
    \vspace{-0.5cm}
      \label{fig:declutter_example}
\end{figure}

Reconstructing a structured 3D scene from a single RGB image is one of the key remaining challenges in computer vision and graphics with applications in content creation, image editing, augmented and virtual reality, and robotic perception \cite{agarwal2024scenecompleteopenworld3dscene}.
These applications often require more than an accurate reconstruction of the \emph{observed} portions of the image --
it also need semantic decomposition of the entire scene into objects and a complete geometry of each object, including occluded regions~\cite{ozguroglu2024pix2gestalt}.
Due to the inherent ambiguity in this task, many existing methods rely on data-driven priors, leveraging pretrained models for semantic segmentation, monocular depth estimation,
and object geometry priors learned from online repositories of CAD models~\cite{zhou2024deep,wu2025dioramaunleashingzeroshotsingleview,Yan:2023:PSDR-Room}.
We found, however, that many image analysis tools fail on complex, cluttered real-world environments with self- and inter-object occlusions, which often leads to compounded errors and poor object completion during reconstruction  (see Fig.~\ref{fig:declutter_example}A). 

To tackle this challenge, we rely on a trivial observation that it is easier to detect, segment, and reconstruct an object if it is not occluded by scene clutter (see Fig.~\ref{fig:declutter_example}B).
Thus, we propose a pipeline for iteratively detecting, segmenting, reconstructing, and removing foreground objects, inpainting the occluded regions, and repeating the process until no objects are left in the scene (see Fig.~\ref{fig:teaser}).
This process offers three advantages.
First, detecting and segmenting a single foreground object is expected to be easier since it will have fewer occlusions.
Second, we leverage powerful 2D generative models to inpaint segmented regions, effectively completing occluded areas of cluttered objects. This leads to cleaner and more complete segmentations — rather than partial ones — of the occluded objects. These complete 2D segmentations provide higher-quality input to 3D reconstruction modules, resulting in more accurate 3D models.
Finally, by relying on off-the-shelf models for more fundamental applications, our method can immediately benefit from advances made in each
of those problems without needing to retrain any component.

The are a couple of technical challenges in designing our pipeline.
Perhaps the most important one is defining the appropriate foreground object for removal.
While this problem seems deceptively simple at first glance, in practice, there are many spatial, physical, and semantic considerations.
For example, removing the closest object could lead to selecting something that is still heavily occluded (e.g., a floor), or removing an 
object that serves as a support can mistakenly lead an inpainting method to re-generate the object instead of an empty region.
Our insight is to leverage a VLM that implicitly contains rich semantic and spatial understanding of natural scenes, and thus can be effectively used to orchestrate the object removal sequencing. 
Another challenge arises while fitting reconstructed objects into a single coherent scene.
A common strategy is to employ monocular depth estimation as a target for object placement procedures.
This approach, however, is not able to handle occluded content.
Our method, however, yields a sequence of progressively decluttered RGB images.
This means we never actually have to deal with occlusions -- there is always at least one image where the object being positioned
is fully visible.
Unfortunately, monocular depth estimations from the decluttered images are often completely incongruent.
We address this problem by devising a depth alignment optimization that allows us to always fit objects against
their unoccluded counterparts while still yielding a single coherent scene.

Our full pipeline is divided in two stages: iterative object removal (stage 1) and layout optimization (stage 2).
The goal of the first stage is to create a sequence of RGB images that depict the scene with a decreasing number of objects.
It starts with an input image, which is fed into a VLM~\cite{openai2024gpt4ocard} with custom prompts
to subsequently yield the name of a candidate foreground object.
We then feed the VLM textual output along with the image into a text-conditioned image segmentation model,
Grounded SAM~\cite{ren2024grounded,sam_hq}, which generates an object mask.
The masked object is then removed via conditional image generation~\cite{kontext,sun2025attentive}.
The whole process is then iteratively repeated with the resulting image until the VLM no longer identifies any objects in the image.
We then move on to stage 2.
We start by feeding the object masks and images computed in the previous stage to a single-view reconstruction model~\cite{hunyuan}.
Then, we run monocular depth estimation on each image~\cite{wang2024moge} and optimize their output
using a depth alignment procedure to consolidate all depth estimates into a coherent space.
Finally, we use geometric fitting aided by VGGT~\cite{vggt} to position and orient the reconstructed complete objects following the consolidated depths to 
create the final scene. 

We evaluate our method on well-known datasets against competitive 3D reconstruction baselines.
When evaluated on the 3D-Front dataset, our method improves the current state-of-the-art with increased performance over F-Score and Chamfer Distance respectively.
Our qualitative results show that, even relying solely on off-the-shelf models without performing any additional fine-tuning,
our method is capable of reconstructing 3D scenes from complex real image inputs, generalizing to domains beyond
the traditional furniture-centric benchmarks.
We also show that our iterative object removal approach can boost segmentation of cluttered objects. On the ADE20K dataset, we compare our approach with object removal against a  single-step  VLM + GroundedSAM baseline, and report an increase in IoU and RandIdx.

Our contributions are summarized as follows.
\begin{enumerate}
\item We introduce a novel VLM "orchestrator" framework that chooses the next object to remove in the scene, and delegates the task to off-the-shelf vision models that yield 3D object reconstruction;
\item We propose a 3D layout optimization method relying on a depth-alignment procedure that consolidates depth estimates of images depicting the same scene with a varying number of objects; and
\item We demonstrate that removing foreground objects helps segmentation of previously occluded objects, and their associated 3D generation.
\end{enumerate}

\section{Related Work}


\textbf{3D Scene Understanding.} Vision-language models (VLMs) have recently demonstrated strong capabilities in open-vocabulary reasoning and spatial understanding \cite{Gu2021OpenvocabularyOD,Chen_2024_CVPR}. Through large-scale vision-language pretraining, these models can describe object relationships and scene structure in ways that suggest an implicit 3D awareness — reasoning about depth, layout, and occlusion even in complex environments~\cite{cheng2024spatialrgpt}. Complementing this, visual foundation models such as SAM~\cite{kirillov2023segany}, Grounding DINO~\cite{liu2023grounding}, and inpainting models like LaMa~\cite{suvorov2021resolution} and Stable Diffusion variants~\cite{Rombach_2022_CVPR} have enabled accurate object segmentation, detection, and removal. However, these visual models alone often struggle to produce coherent results in the presence of occlusions or complex interactions between objects. Our approach combines both types of models by using VLMs not only as semantic descriptors but also as structural guides, orchestrating a unified, training-free pipeline that integrates visual foundation models for segmentation and inpainting to robustly decompose and reconstruct layered 3D layouts. \\


\noindent \textbf{Image-to-3D.} Diffusion/NeRF-based approaches have made significant strides in generating 3D assets from images or text. Many recent works leverage Score Distillation Sampling~\cite{poole2022dreamfusion}  to guide implicit representations (such as NERFs or Gaussian Splats) using powerful 2D diffusion priors~\cite{Rombach_2022_CVPR}. These methods, such as DreamFusion, Magic3D~\cite{lin2023magic3d}, produce high-fidelity results at the object level, but typically assume sparse scenes or centered objects. When applied to full scenes, they often struggle with occlusion, spatial consistency, and scalability, and tend to generate entangled or non-editable outputs~\cite{10.5555/3692070.3692570}. Our method takes a different path: instead of distilling diffusion priors into scene geometry, we use foundation models directly for segmentation, removal, and reassembly of objects in cluttered environments, yielding explicit, editable, and semantically consistent scene layouts. \\


\noindent \textbf{Scene Generation with LLMs.} There has also been a growing body of work in using Large Language Models (LLMs) to generate structured scene representations, often combining language priors with symbolic scene graphs or layout templates~\cite{10.5555/3666122.3666924,10.5555/3692070.3692846, gumin2025imperativevsdeclarativeprogramming}. These systems can generate impressive results in constrained domains, such as indoor scenes with canonical room structures or familiar object arrangements~\cite{fu2024anyhome}. However, they often struggle to generalize beyond the distribution of their training data and often rely on static architectural priors such as room boundaries or floor plans~\cite{littlefair2025flairgpt, aguinakang2024openuniverse}. In contrast, our system bypasses the need for symbolic reasoning by operating directly on image input and leveraging foundation models that generalize naturally to a broad set of scenes, including outdoor, cluttered, or atypical environments, without requiring handcrafted structural priors. \\

\noindent \textbf{Single-view scene reconstruction.} Inferring a complete 3D scene from a single RGB image remains especially challenging due to severe occlusions, weak depth cues, and complex object interactions. Early approaches such as DeepPriorAssembly~\cite{zhou2024deep} and PSDR-Room~\cite{Yan:2023:PSDR-Room} assume a fixed, closed set of object categories and exploit strong architectural priors in structured indoor environments, which constrains their ability to generalize beyond those domains. More recently, CAST~\cite{yao2025castcomponentaligned3dscene} introduces an occlusion-aware, object-centric pipeline: each object is generated independently by retraining large-scale 3D generative models, followed by a physics-aware fitting scheme to resolve inter-object collisions. While CAST impressively tackles occlusions and physical consistency, it requires costly retraining and task-specific model adaptations. By contrast, our pipeline is fundamentally powered by a pretrained VLM~\cite{openai2024gpt4ocard} that drives every decision—reasoning about hidden geometry, occlusions, and physical plausibility—within a fully training-free, modular framework: the VLM orchestrates off-the-shelf segmentation to partition the scene into layers and then guides sequential object removal to reconstruct each layer. Centering the VLM as the core decision engine yields a scalable, editable scene synthesis pipeline that operates without supervision, architectural priors, or bespoke training.


\begin{figure}[t]
  \centering
  \includegraphics[width=\columnwidth]{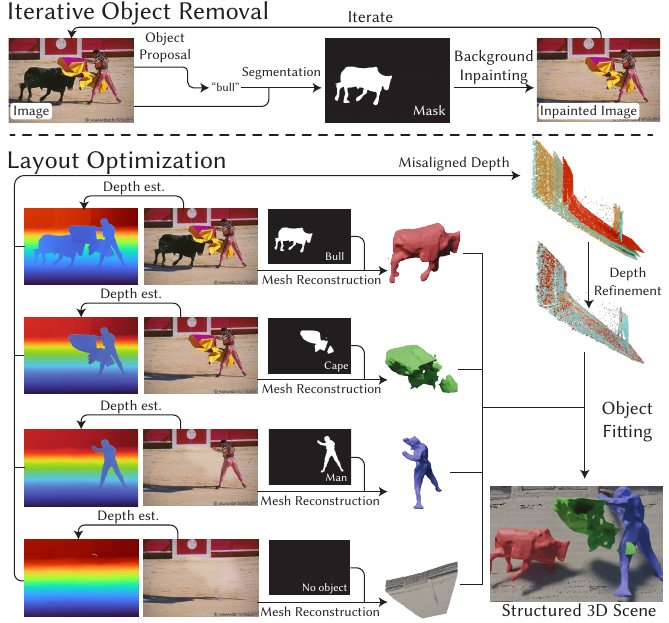}
  \caption{
  \textbf{Overview}.
  Our method consists of two stages. In the first, iterative object-removal stage, we use custom VLM prompting to identify the next best candidate for removal. Based on the predicted object name, a segmentation module computes its mask, and an inpainting module fills in the background. In the second, layout-optimization stage, we reconstruct a complete mesh for each removed object using the sequence of images and their corresponding masks. To place the objects into a shared scene, we first apply monocular depth estimation, then perform depth-map refinement to align all independently predicted depth maps. Finally, we translate and scale each object to fit the unified point cloud, producing the final 3D scene.
  }
  \label{fig:pipeline}
\end{figure}

\section{Method}
Our method consists of two stages: Iterative Object Removal and Layout Optimization.
The first stage iteratively removes objects from the input image until only the background remains.
The output of this stage is a sequence of (gradually emptier) images and masks indicating which objects are removed at each iteration. 
The second stage starts by feeding those image-mask pairs into a single-view reconstruction model~\cite{wang2024crm} to create
a single 3D shape per pair (except for the last pair that should not contain any object, just a background).
Additionally, for every RGB image in the sequence we also estimate its depth using a monocular depth estimation network~\cite{wang2024moge}.
Finally, we optimize the depth estimations to enforce their congruency and fit every object to their respective (refined) depth estimation.
Details on these stages are described below.

\subsection{Iterative Object Removal}
We adopt an iterative, VLM-guided inpainting procedure to decompose a given RGB image $\text{I}$ into a sequence of \(N\) images $(\text{I}_n)^N_1$ , where each image $\text{I}_n$ is obtained by removing an object from $\text{I}_{n-1}$, and $\text{I}_1 = \text{I}$.
The goal of this process is to generate amodal segmentation masks $(\text{M}_n)^N_1$ for all objects present within the scene.
Amodal masks contain both the visible and occluded parts of an object.
Specifically, at each step, we prompt the VLM to identify a main object along with its corresponding supported objects,
which are then removed from the image.
A graphical overview of this stage is presented on the top section of Figure~\ref{fig:pipeline}. \\

\noindent \textbf{Object Proposal.} Given a single RGB image, our approach uses a VLM as an orchestrator to systematically disentangle the complex scene into a sequence of objects.
Initially, the VLM is instructed to pick the \texttt{closest and fully distinct visible object without any occlusions} (unless it is at the edge of the camera).
The output of the VLM is an object name like \texttt{human}, \texttt{chair}, or \texttt{vase}.
The VLM is also instructed to note any objects positioned on top or contained within the main object--such as plates on a table or books on a bookshelf, and to provide a textual description of the object.
We refer to those as \emph{secondary objects}.
When secondary objects are found, they are handled first and the main object is postponed to the next iteration.
If secondary objects are found to have secondary objects of their own (e.g. on a table there is a vase that has plants), the VLM is instructed to create a label representing both (e.g. \texttt{a vase with plants}).
If the VLM no longer identifies any objects within the scene, the inpainting process stops.
The full prompt in provided in the supplementary material.

\noindent \textit{Implementation.} We use ChatGPT-4o as our VLM orchestrator~\cite{openai2024gpt4ocard}. \\

\noindent \textbf{Object Segmentation.}
At an iteration $n$ of the object removal process, after obtaining an object name from the VLM (object proposal step), the next step is to segment them in the given image $I_{n}$ yielding a mask $M_n$.
We do that by using a Grounded Segment-Anything model (Grounded-SAM)~\cite{ren2024grounded}, equipped with High-Quality SAM (HQ SAM)~\cite{sam_hq}, to detect precise segmentation masks conditioned on the text descriptions provided by the object proposal. 
It is important to notice that we always segment an object free of occlusion since we query the VLM for the
\texttt{closest and fully distinct visible object without any occlusions}.
Such object is much easier to be segmented than an arbitrary one, and, because we always remove it after segmentation (see next step), there will always be an object free of occlusions for as long as there are objects in the scene. \\

\noindent \textbf{Foreground Removal.}
The last step of the $n$-th iteration consists of removing the segmented object from the RGB image.
The result of this step is an image $I_{n+1}$ containing one less object.
We investigate two ways of accomplishing this task.
The first one involves using Flux Kontext~\cite{kontext} with an input image outlining the object to be removed followed
by an object removal instruction.
Specific details can be found in supplemental material.
The second one employs a much smaller and less general image inpainting model.
In this approach, to avoid possible pitfalls from incorrect segmentations, each segmentation mask is dilated slightly before it is passed to an object-removal model~\cite{sun2025attentive}.
Existing object-removal inpainting models often produce artifacts influenced by the context of the surrounding image.
To avoid this, we further dilate the mask used for object-removal and crop the image to the bounds of this dilated mask.
We query the VLM to generate a prompt for a second, promptable inpainting model~\cite{sdxl} by requesting positive terms associated with background materials (e.g. wooden flooring, white walls, greenery, etc.) and negative terms associated with objects that could fit within the mask, but should not be added to the image.
The application of this second model finally produces the next image in the sequence $\text{I}_{n+1}$.

All these steps are  repeated until the VLM no longer identifies objects in the current inpainted image or when there are no objects detected by Grounded SAM.
The result of this procedure is a sequence of amodal masks $(\text{M}_n)^N_1$ and a sequence of RGB images depicting a progressively emptier
scene $(\text{I}_n)^N_1$, where $I_1$ corresponds to the full image provided by the user and $I_n$ is an image without any objects, just
a background.

\subsection{Layout Optimization}
\label{subsec:layout}
After the first stage is complete, we are given two sequences of $(M_n)^N_1$ and $(I_n)^N_1$ corresponding to binary masks and RGB images,
respectively, computed during the last stage.
An important relationship to keep in mind is that a mask $M_n$ contains the the object to be removed from $I_n$, and $I_{n+1}$ is the
result of removing $M_n$ from $I_n$
An illustration of this stage can be found on the bottom section of Figure~\ref{fig:pipeline}. \\

\noindent \textbf{Mesh Reconstruction.} Following the generation of the sequence of masks $(\text{M}_n)^N_1$ and unpainted images $(\text{I}_n)^N_1$, we pass each masked object, \ie  $(\text{M}_n \odot \text{I}_n)^N_1$
where $\odot$ is the Hadamard product, into the Hunyuan2~\cite{hunyuan}, an off-the-shelf image-to-3D model, to generate a sequence of textured meshes $(\mathcal{M}_n)^{N-1}_1$.
Note that in the original images, some of these objects can be occluded, making it impossible to directly pass them to an image-to-3D generator without obtaining degenerate results.
Our object removal strategy is critical to obtain objects and their masks free of any occlusions -- this leads to significantly better
object reconstructions. 
However, generated meshes do not adhere to the in situ pose of the target object.
We make use of a 3D fitting procedure that translates and scales the generated meshes to better align with a set of depth maps estimated by MoGe~\cite{wang2024moge}.
The depth estimator is applied to each image in the sequence $(\text{I}_n)^N_1$, which gives us a sequence of disparity maps $({{\mathcal{D}}}_n)^N_1$.
Unfortunately, because the estimation are computed independently, their disparities are often dramatically different even
in regions of the image that are always fully visible. \\

\begin{figure}
      \centering
      \includegraphics[width=\linewidth]{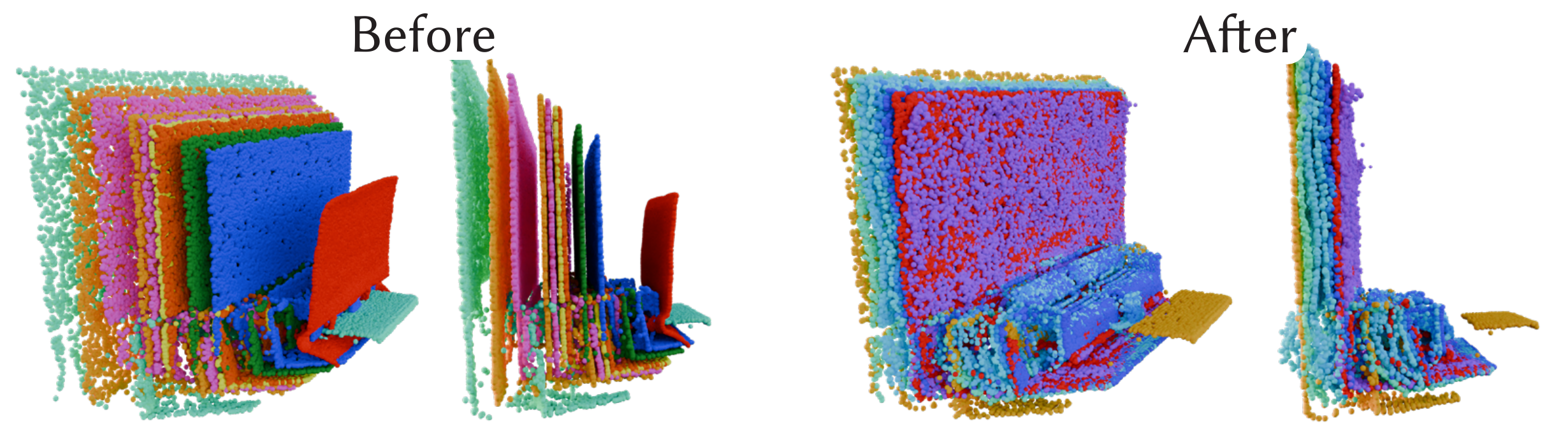}
      \vspace{-0.4cm}
      \caption{\textbf{Depth refinement}.
      When each object-removed image is passed independently through a standard depth estimator, the resulting depth maps do not align. To reconcile these discrepancies, we introduce a depth‐refinement optimization that jointly adjusts the estimated depths into a single, coherent representation. Left: Point clouds reconstructed from the raw depth estimates, color‐coded by their source image. Right: the same scene after applying our alignment procedure.  Notice how the optimization brings all point clouds into tight agreement.}
      \vspace{-0.4cm}
      \label{fig:depth}
    \end{figure}

\noindent \textbf{Depth map refinement.}
Despite a common background visible in all layers, monocular depth estimator models produce inaccuracies unique to each layer.
To correct disparity misalignments across depth maps, we introduce the following coordinate-based MLP
$f_{\theta_n}$
, that refines the sequence of disparity maps into $({{\mathcal{D}'}}_n)^N_1$ such that:
\begin{equation}
\scriptsize
\mathcal{D}'_n(x,y) \;=\;
\begin{cases}
\mathcal{D}_n(x,y) & \text{if } n = 1, \\
\begin{aligned}
f_{\theta_n}\bigl(\tilde{x}, \tilde{y},\mathcal{D}_n(x,y),\text{I}_n(x,y)\bigr),
\end{aligned} & \text{for } n = 2, \ldots, N.
\end{cases}
\end{equation}
where, $\tilde{x}$ and $\tilde{y}$ are normalized pixel coordinates, and $\mathcal{D}_1$ serves as a fixed reference and is not refined.  This refinement method is similar to that of~\cite{kopf2021rcvd}, but our MLP-based approach offers greater flexibility compared to their bilinear spline-based method.
To encourage consistency across adjacent layers, we define a loss that penalizes disparity differences in non-occluded regions:
\begin{equation}
  \mathcal L(\theta)
  = \sum_{n=1}^{N-1} \sum_{x,y}
    \bigl(1 - M_n(x,y)\bigr)\,
    \bigl\lvert \mathcal{D}'_n(x,y) - \mathcal{D}'_{n+1}(x,y)\bigr\rvert.
\end{equation}
We model each $f_{\theta_n}$ as a small 3-layer perceptron and optimize \(\mathcal L\) over \(\theta_2,\dots,\theta_{L}\) using the Adam optimizer.
The effect of performing the depth refinement is illustrated in Figure~\ref{fig:depth}. \\

\noindent \textbf{Object Fitting.} 
The goal is to estimate a similarity transform $T \in \mathrm{Sim}(3)$ that
maps the mesh coordinate frame of a 3D mesh $\mathcal{M}$ to the
image/disparity space of a segmented input image $\text{I}$ with object mask
$\text{M}$ and disparity map $\mathcal{D}$ (aligned with $\text{I}$).
The procedure begins by rendering $\mathcal{M}$ under $S$ yaw offsets,
producing a collection of synthetic views $\{\mathcal{R}_i\}_{i=1}^S$. These are passed
together with the segmentation-masked image $\text{I}$ into VGGT~\cite{vggt}, which predicts a
coarse rotation estimate $R_{est}$. The mesh is then rendered again with this
rotation removed, yielding an image $\text{I}_{rot}$ and a corresponding disparity map
$\mathcal{D}_{rot}$ in a canonical pose.

While the first stage to our alignment procedure accounts for gross misalignment between source image and generated mesh, we wish to further refine the alignment.  VGGT is applied once more to the
pair $(\text{I}, \text{I}_{rot})$, producing a set of candidate 2D
correspondences. Only correspondences with
confidence above a threshold $K$ and within the mask $\text{M}$ are retained.
If the number of valid correspondences exceeds a minimum $N_{\min}$, the
associated image points are backprojected into 3D using their disparity
values, and the similarity transform $T$ is computed by solving a
least-squares alignment problem between the two resulting point sets. If
instead the number of reliable matches is too low, a fallback strategy is
used: both $\mathcal{D}$ (restricted to $\text{M}$) and $\mathcal{D}_{rot}$
(rendered over the full field of view) are converted into point clouds, and
$T$ is estimated via an Iterative Closest Point (ICP) procedure. This two-stage design ensures
robustness, using direct correspondence fitting when possible and defaulting
to dense disparity alignment when correspondences are sparse.

The intuition behind this procedure is to decouple the alignment problem into
two stages. In the first stage, the goal is to determine a coarse rotation
estimate $R_{est}$ that orients the mesh $\mathcal{M}$ so that its pose matches the
appearance of the cropped object in $\text{I}$. This coarse alignment is
sufficient to place the mesh in the correct orientation but does not yet
resolve its position within the full scene. In the second stage, the method
refines the alignment by exploiting correspondences between
$\text{I}$ and $\text{I}_{rot}$. This step incorporates the disparity
information $\mathcal{D}$ defined over the full image, thereby grounding the
mesh placement with respect to the global camera intrinsics and scale. In
other words, while the cropped object provides local orientation cues, only
the full image supplies the geometric context required to determine the
absolute position of $\mathcal{M}$ in the scene. This separation of rotation
estimation and full-scene placement makes the fitting procedure both stable
and robust. The full algorithmic details are summarized in Algorithm~\ref{alg:fit-object}.

\begin{algorithm}[t]
\caption{Object Fitting via Segmentation–Render Correspondence}
\label{alg:fit-object}
\KwIn{$\text{I}$: segmented image with mask $\text{M}$ \\
\hspace{1.7em}$\mathcal{D}$: disparity aligned with $\text{I}$ (after refinement) \\
\hspace{1.7em}$\mathcal{M}$: 3D mesh \\
\hspace{1.7em}$K$: minimum confidence for correspondences \\
\hspace{1.7em}$S$: number of yaw renders (e.g., $8$)}
\KwOut{$T$: $\mathrm{Sim}(3)$ transform mapping mesh space to image/disparity space}

Render $S$ yaw views of mesh $\mathcal{M}$: $\{\mathcal{R}_i\}_{i=1}^S$\;

$R_{\mathrm{est}} \gets \mathrm{VGGT.EstimateRotation}(\text{I}, \{\mathcal{R}_i\})$\;

$\text{I}_{\mathrm{rot}},\, \mathcal{D}_{\mathrm{rot}} \gets \mathrm{Render}(\mathcal{M}; R_{\mathrm{est}})$\;

$(xy_{\text{I}},\, xy_{\mathrm{rot}},\, \mathrm{conf}) \gets \mathrm{VGGT.Track}(\text{I}, \text{I}_{\mathrm{rot}})$\;

Filter correspondences:
$\mathrm{idx} \gets \{\, j \mid \mathrm{conf}[j] \ge K,\ xy_{\text{I}}[j] \in \text{M}\,\}$\;

\eIf{$|\mathrm{idx}| \ge N_{\min}$}{
    $X_A \gets \mathrm{Backproject}(\mathcal{D},\, xy_{\text{I}}[\mathrm{idx}])$\;
    $X_B \gets \mathrm{Backproject}(\mathcal{D}_{\mathrm{rot}},\, xy_{\mathrm{rot}}[\mathrm{idx}])$\;
    $T \gets \mathbf{LEAST\_SQUARES}(X_B, X_A)$\;
}{
    $P_A \gets \mathrm{DisparityToPointCloud}(\mathcal{D},\, \text{M})$\;
    $P_B \gets \mathrm{DisparityToPointCloud}(\mathcal{D}_{\mathrm{rot}},\, \text{all},\, \mathrm{FOV})$\;
    $T \gets \mathbf{ICP}(P_B, P_A)$\;
}
\Return $T$\;
\end{algorithm}
In the end, after applying the fitting procedure to all meshes, we get a sequence of aligned meshes $(\mathcal{M'}_n)_1^{N-1}$ which constitutes a structured reconstruction of the input image $\text{I}$, where each object is cleanly separated and reconstructed individually.
Notably, the last image $I_N$ in the sequence created during the first stage does not contain any object -- it only has a background.
For this specific case, we create a mesh following the refined depth $\mathcal{D}^\prime_N$ by simply tessellating the adjacent pixels;
\ie for every pixel coordinate $x, y$ create a triangle connecting the points in $\langle x, y\rangle$, $\langle x+1, y\rangle$, $\langle x, y+1 \rangle$ and $\langle x+1, y \rangle$, $\langle x+1, y+1 \rangle$, $\langle x, y+1 \rangle$.
In all figures throughout this paper, the background mesh is colored in gray.

\noindent \textit{Object Filtering.} Once all objects have been fitted to reconstruct the full scene, we remove any object whose volume overlaps more than 90\% with another object generated earlier in the sequence (i.e. if object A was fitted before object B and their Intersection-over-Union is greater than 0.9, we discard B). This step addresses rare cases in which the inpainting model mistakenly inserts a different object into the scene—for example, replacing one table with another.

\section{Results}


\textbf{Datasets.} We quantitatively evaluate our method on two standard benchmarks: the 3D‑Front \cite{Fu2021FRONT3D} dataset for quantitative assessment of 3D reconstruction quality, and the ADE20K \cite{zhou2017scene} dataset to measure scene segmentation performance.

3D‑Front \cite{Fu2021FRONT3D} is a large‑scale, fully synthetic dataset of indoor environments composed of professionally modeled room layouts. To conduct quantitative comparisons, we sample 600 images and associated ground truth 3D reconstructions from the preprocessed by PanoRecon~\cite{dahnert2022panorecon} to form our single‑image dataset. All images are rendered with the virtual camera held level to the ground plane at a height of 0.75m above the floor. For each image, the ground‑truth mesh is obtained by keeping only the geometry within the camera’s view frustum and discarding everything else. 

The ADE20K \cite{zhou2017scene} dataset is the basis for the MIT Scene Parsing Benchmark \cite{zhou2016semantic}, which provides a standard training and evaluation platform for scene parsing algorithms. The ADE20K dataset contains over 20,000 scene‑centric images exhaustively annotated with objects and object parts across 150 semantic categories, including “stuff” classes (e.g., sky, road, grass) and discrete object classes (e.g., person, car, bed). We sample 500 scenes from the test set.

\begin{figure}
  \centering
  \includegraphics[width=\linewidth]{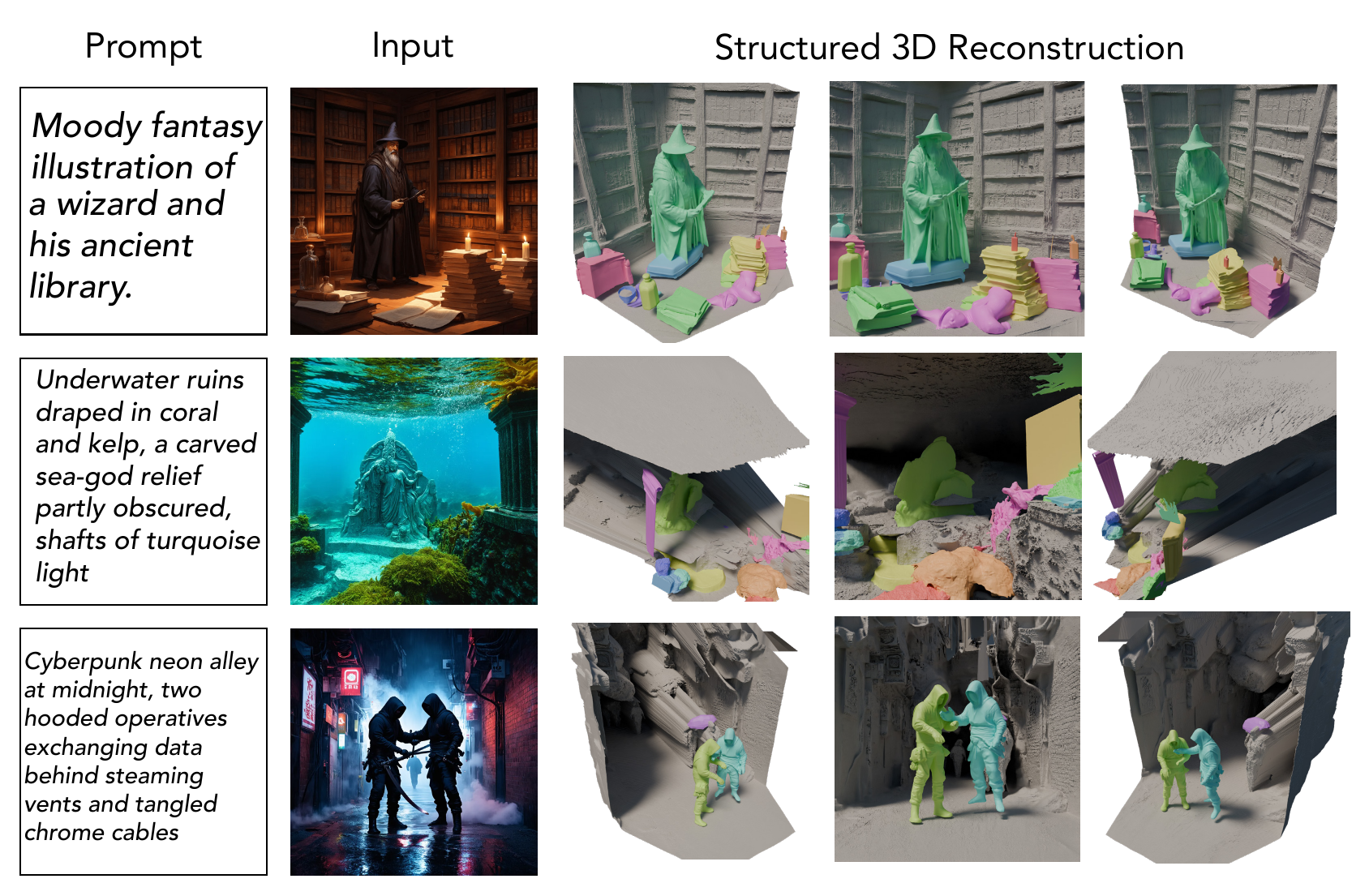}
  \vspace{-8pt}
  \caption{
  \textbf{Text-to-3D Scene pipeline.}
  We show that our method can be applied to images generated by text-to-image models~\cite{sdxl}.
  This corresponds to a fully automated text-to-3D scene pipeline.
  }
  \label{fig:textto3D}
\end{figure}

\begin{figure}
  \centering
  \includegraphics[width=\linewidth]{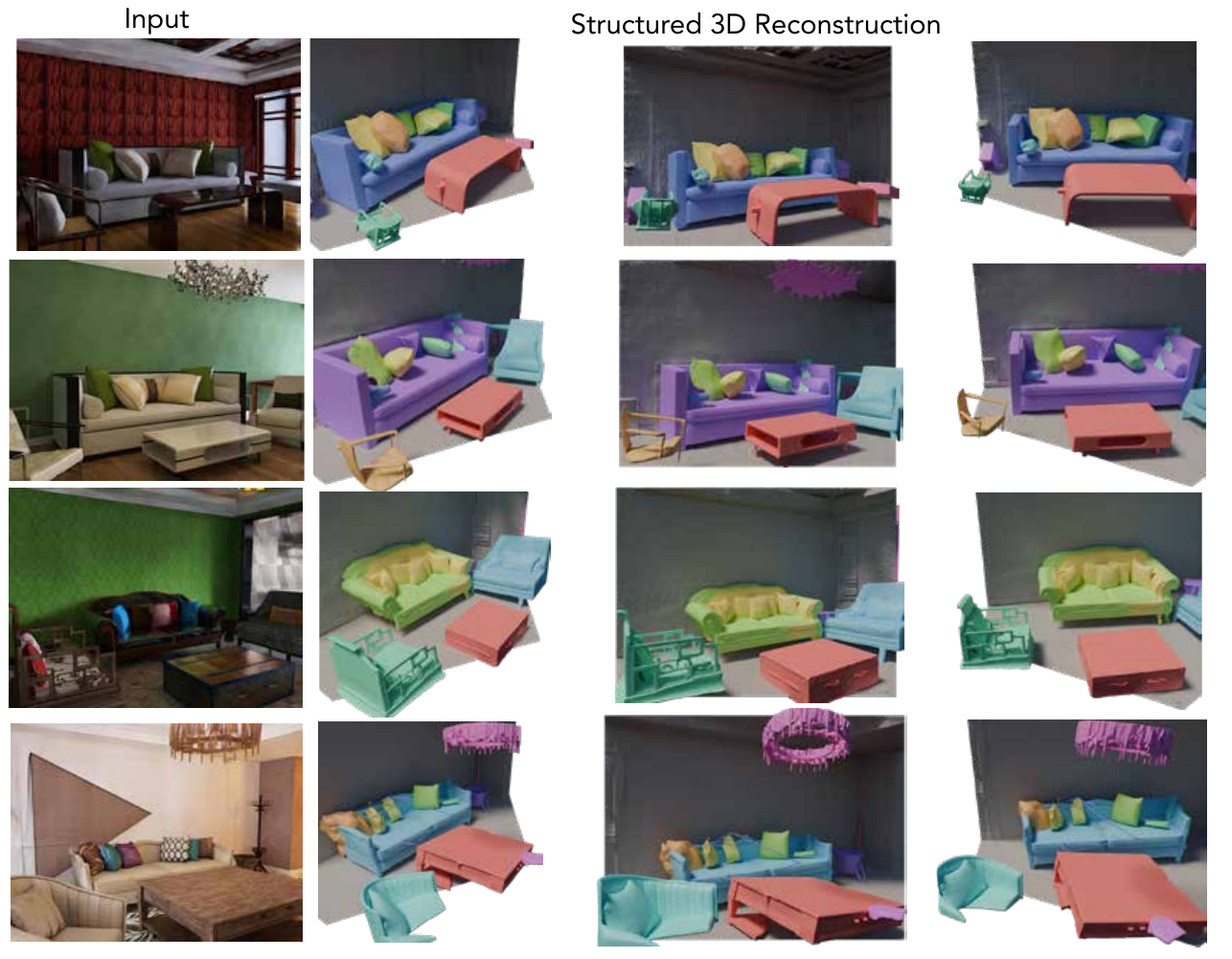}
  \caption{\textbf{In-the-wild indoor scenes.}
  Our method is capable of generating 3D reconstructions of individual objects and backgrounds without being restricted to specific categories.
  }
  \label{fig:qualitative-indoor}
\end{figure}

\begin{figure*}
      \centering
      \includegraphics[width=\linewidth]{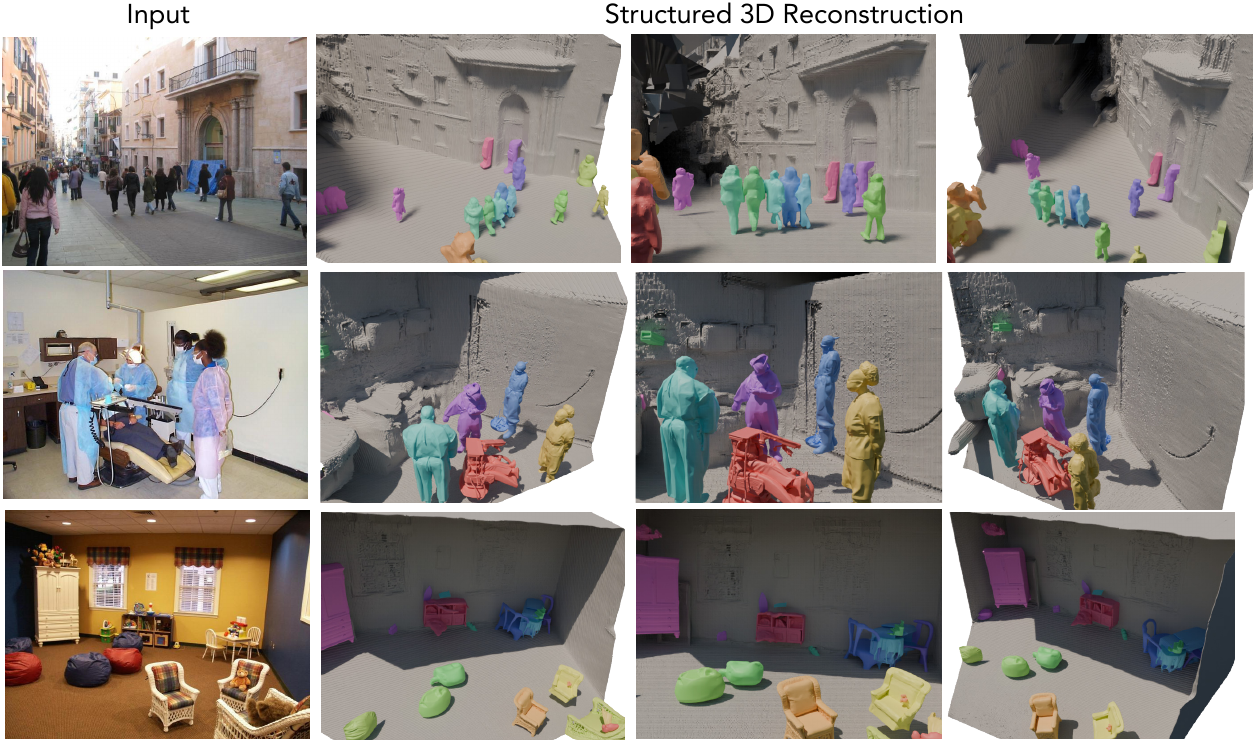}
      \caption{A gallery of qualitative results on in-the-wild indoor and outdoor scenes.}
      \label{fig:qualitative-outdoor}
\end{figure*}

Qualitative results in outdoor and indoor scenes can be found in Figures~\ref{fig:qualitative-outdoor} and~\ref{fig:qualitative-indoor},
respectively.
We also demonstrate the performance of our method when applied to synthetically generated images in Figure~\ref{fig:textto3D}.
This setup comprises a fully automatic text-to-3D pipeline.

\subsection{Reconstruction Evaluation on 3D-Front}


\textbf{Baselines.} We compare our method against leading single-view 3D reconstruction approaches.
\textit{Gen3DSR}~\cite{gen3dsr} and \textit{MIDI*}~\cite{midi} leverage pre-trained open entity segmentation systems, depth, and object reconstruction models.
Differently from our approach, they deal with occluded objects by training occlusion-aware models that will operate in individual scene objects.
Gen3DSR proposes training an image-based amodal completion model while MIDI creates an amodal object reconstruction system.
Since the original MIDI predictions rely on predetermined class labels, we estimated those using VLMs and used that in our evaluation (MIDI*).
Qualitative comparisons can be found in the supplemental material.
For ablations, we assess two key variations of our pipeline: one without the object filtering step in object fitting stage (to evaluate the impact of object removal artifacts), and another without depth alignment, where each object is fitted using its raw inpainted depth map; see Section~\ref{subsec:layout} for details.
Full results on these can be found in the supplemental material\\

\noindent \textbf{Metrics.} We evaluate reconstruction quality using several standard metrics: Chamfer Distance (CD), F-Score (FS), Object-level F-Score (Obj-FS), Depth Error, Segmentation IoU, and Mesh IoU (M-IoU). Following~\cite{Peng2021SAP}, we use a 0.1 ($\approx10cm$) threshold for CD and FS on 3D-FRONT dataset, and sample 10k points per object for geometric comparisons. Obj-FS is computed by matching ground-truth objects with their best predicted object pair. Depth error is measured as the mean absolute difference in depth renderings. Segmentation IoU measures the alignment between predicted 2D masks and ground-truth labels, while Mesh IoU evaluates segmentation rendered from the reconstructed mesh. Note that our method fully reconstructs the scene background using the depth given by the last iteration of inpainting, but in these evaluations, only the objects are considered for metric computations. \\

\noindent \textbf{Quantitative Results.} As shown in Table~\ref{tab:front3d_metrics}, our method outperforms all baselines across every metric, including Chamfer Distance, F-Score, Depth, and Segmentation. Compared to systems like Gen3DSR and MIDI*, our approach yields higher object-level fidelity and geometric accuracy. Interestingly, the performance of our model remains robust even without object filtering, suggesting that artifacts introduced during object removal have limited adverse effect on final reconstruction. In contrast, disabling depth alignment leads to consistent degradation across all metrics, underscoring the value of maintaining coherent depth priors throughout the reconstruction pipeline. Qualitative comparisons against other methods~\cite{Gkioxari2019mrcnn,zhou2024deep}
can be found in the supplemental material. \\

\noindent \textbf{Robustness to Scene Complexity.} We evaluate the robustness of our method when exposed to varying levels of scene complexit in the
supplemental material.

\begin{table}[ht]
\centering
\resizebox{\linewidth}{!}{%
  \begin{tabular}{lcccc}
    \toprule
    Model & Chamfer $\downarrow$ & P@0.1 $\uparrow$ & R@0.1 $\uparrow$ & F1@0.1 $\uparrow$ \\
    \midrule
    Gen3DSR                           & 0.12 & 67.54 & 74.50 & 70.18 \\
    Gen3DSR (w/ backgrounds)          & 0.21 & 59.11 & 53.31 & 55.48 \\
    MIDI*                             & 0.24 & 51.95 & 48.67 & 49.36 \\
    \midrule
    Ours                              & 0.11 & 72.58 & 73.38 & 71.65 \\
    Ours (w/ backgrounds)             & 0.17 & 72.58 & 55.56 & 61.69 \\
    Ours (filtered)                   & 0.12 & 71.03 & 69.90 & 68.85 \\
    Ours (filtered, w/ backgrounds)   & 0.17 & 70.96 & 54.01 & 60.10 \\
    \bottomrule
  \end{tabular}%
}
\caption{
\textbf{Quantitative evaluation on 3D-Front.}
Our method outperforms the closest baselines (Gen3DSR and MIDI*) across all metrics,
except for recall when background regions are excluded from the evaluation,
where Gen3DSR achieves slightly higher recall.
The results further highlight the importance of depth refinement, which consistently improves
performance across all metrics.
}
\label{tab:front3d_metrics}
\end{table}


\vspace{-10pt}
\subsection{Segmentation Evaluation on ADE20K}

To assess our system’s scene parsing capability in more complex, real-world scenarios, we perform an evaluation on a subset of the ADE20K dataset \cite{zhou2017scene}. Unlike the 3D‑Front dataset, which contains only 1–5 objects per scene and is synthetic in nature, ADE20K comprises richly annotated real-world scenes that often feature a significantly higher number of objects, occlusions, and clutter. Although 3D ground truth is not available for these scenes, we can still benchmark our system’s ability to parse and reconstruct scenes by measuring the segmentation performance.

\begin{figure}
      \centering
      \includegraphics[width=1\linewidth]{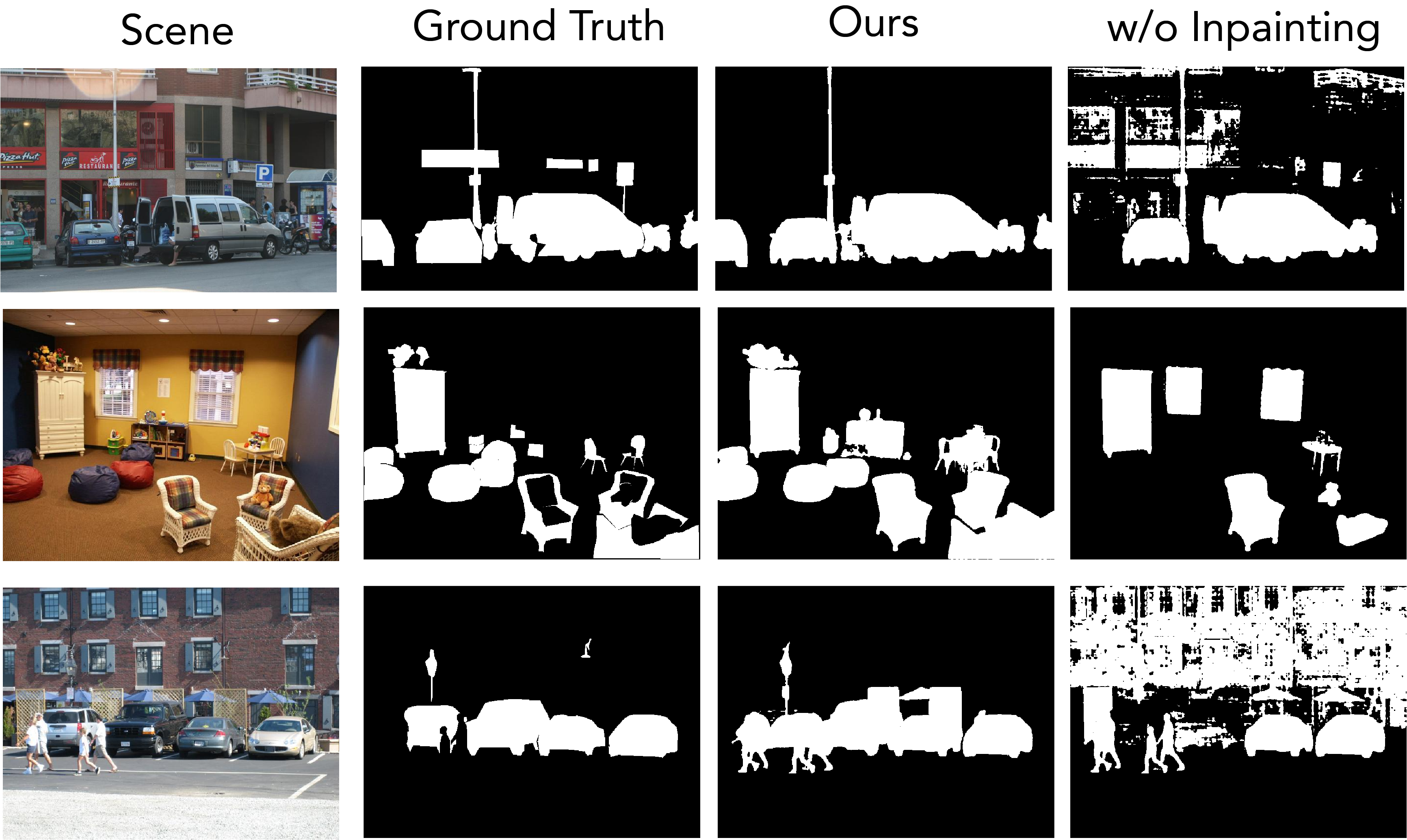}
      \caption{
      \textbf{Qualitative comparison on ADE20K segmentations.}
      From left to right: input, ground truth, with iterative object removal, and without iterative object
      removal.}
      \label{fig:ade-comparison}
\end{figure}

Our evaluation here focuses on testing the efficacy of our object removal pipeline, which plays a critical role in enabling robust object segmentation and reconstruction under heavily occluded or cluttered environments. To isolate the contribution of object removal, we compare our full pipeline with an ablated version that omits object removal—akin to Gen3DSR. We show some examples in Figure~\ref{fig:ade-comparison}. 
We conduct our quantitative evaluation in two settings.
One measures the IoU and RandIdx over \emph{things} classes~\cite{caesar2018cvpr}—i.e., discrete, countable objects such as chairs or lamps—as opposed to \emph{stuff} classes like sky or floor.
Since our system targets object-centric reconstruction, this is the most relevant metric.
The other setting also includes classes from the \emph{stuff} but excludes the ones explicitely prompt the VLM to ignore.
We refer to this subset of classes as \emph{stuff*}.

\noindent \textbf{Quantitative Results.} As shown in Table \ref{tab:segmentation_iou}, our method outperforms the ablation consistently across all evaluation settings. In particular, the largest gain is observed in the \emph{things} category, highlighting the importance of our iterative object removal pipeline for discovering and segmenting occluded or overlapping objects. The \emph{things+stuff} evaluation further validates that our system maintains high precision even under stricter constraints.

\begin{table}[h!]
\centering
\resizebox{\linewidth}{!}{
\begin{tabular}{l@{\hspace{7em}}cc}
\toprule
Method & IoU & RandIdx \\
\multicolumn{3}{c}{
  \makebox[\linewidth]{\hfill\rule{0.5\linewidth}{0.4pt}\hspace{0.5em}%
  \scriptsize Things\hspace{0.5em}\rule{0.5\linewidth}{0.4pt}\hfill}
} \\
Ours (w/ obj. removal - Kontext~\cite{kontext})                    & \textbf{0.51} & 0.29 \\
Ours (w/ obj. removal - inpainting~\cite{sun2025attentive})              & 0.49          & \textbf{0.32} \\
Ours (w/o obj. removal)           & 0.45          & 0.27 \\
Gen3DSR                 & 0.50          & 0.23 \\
\multicolumn{3}{c}{
  \makebox[\linewidth]{\hfill\rule{0.45\linewidth}{0.4pt}\hspace{0.5em}%
  \scriptsize Things \& Stuff \hspace{0.5em}\rule{0.45\linewidth}{0.4pt}\hfill}
} \\
Ours (w/ obj. removal - Kontext~\cite{kontext})                   & \textbf{0.44}          & 0.27 \\
Ours (w/ obj. removal - inpainting~\cite{sun2025attentive})              & \textbf{0.44} & \textbf{0.29} \\
Ours (w/o obj. removal)           & 0.35          & 0.20 \\
Gen3DSR                 & 0.41          & 0.16 \\
\bottomrule
\end{tabular}
}
\caption{Segmentation IoU scores on the SceneParsing dataset.
\emph{stuff*} corresponds to classes belonging to the stuff segment, but \emph{excluding} the ones we specifically asked to be ignored
in our prompt.
“Ours” designates our original pipeline; “w/o obj. removal” is an ablation that performs segmentation in a single step without the our iterative object removal procedure.\label{tab:segmentation_iou}}
\end{table}


\paragraph{Limitations \& Future work}

While our method shows strong performance across both synthetic and real-world scenes, several limitations remain.
We consider the object alignment to be the most challenging portion of the current pipeline.
Even though VGGT provides reasonable estimates in most of the cases, there are scenarios where the initial rotation estimation
is imprecise and insufficient reliable correspondences are found.
The object removal system is also not perfect -- inpainting-based~\cite{sun2025attentive} approach can sometimes lead to artifacts and ghost objects while
text-guided editing alternatives~\cite{kontext} sometimes fail to remove any object when the category provided by the VLM is not precise.
Future work could explore the use of video depth estimation models to better leverage the sequence of images produced during object removal. As foundation models and generative systems continue to improve, our modular pipeline is well positioned to take advantage of these advances.



\section{Conclusion}
\vspace{-5pt}
We introduce \textit{SeeingThroughClutter}, a training-free framework for open-vocabulary 3D scene understanding from a single RGB image. By leveraging vision-language models (VLMs) as scene orchestrators and combining them with iterative object removal, segmentation, and depth-guided layout optimization, our system decomposes complex real-world scenes into structured 3D layouts. Unlike prior work, which treats scene understanding as a one-shot labeling or segmentation problem, our method reasons over object relationships and occlusions—enabling accurate amodal segmentation and geometry recovery in cluttered environments.
Through experiments on both synthetic (3D-Front) and real-world (ADE20K) benchmarks, we demonstrate that our system not only outperforms existing state-of-the-art methods in 3D reconstruction metrics but also remains robust under increasing scene complexity. 

{
    \small
    \bibliographystyle{ieeenat_fullname}
    \bibliography{main}
}




\clearpage
\setcounter{page}{1}
\maketitlesupplementary

\newtcolorbox{promptbox}[1][]{
  colback=gray!10,
  colframe=black,
  boxrule=0.4pt,
  arc=2mm,
  left=6pt,
  right=6pt,
  top=6pt,
  bottom=6pt,
  fonttitle=\bfseries,
  title=VLM Prompt,
  listing only,
  listing engine=listings,
  enhanced,
  breakable,
  width=\linewidth,
  #1,
  before upper=\footnotesize,
  listing options={
    basicstyle=\ttfamily\scriptsize,
    breaklines=true,
    escapeinside=||,
    columns=flexible,
    keepspaces=true,
    upquote=true,
    showstringspaces=false,
  }
}

\title{Supplementary Material for ``Seeing Through Clutter: Structured 3D Scene Reconstruction
via Iterative Object Removal''}

\maketitle
\thispagestyle{plain}

\section{VLM orchestrator}

We provide the prompt we use to query the VLM (GPT-4o) and prompt to remove objects (Flux Kontext).

\subsection{Amodal selection}

\begin{promptbox}[title=Amodal Selection Prompt]
\# Instructions

1. You will be given an image of a scene. First, describe what the scene is.

2. Examine the scene and identify the object that is closest to the camera. This object should be distinct and fully visible, meaning it is not occluded by any other objects (unless part of it is outside the camera's frame). If it is occluding any objects, note which objects it is occluding. If there are no objects in the scene, state that the scene is empty.

3. If the scene is not empty and you have identified the closest fully visible object, explain why it might be there in relation to the full scene. Explain if there are any objects adjacent to the closest fully visible object and explain how they are behind the closest fully visible object. Use your explanations to re-evaluate what the closest fully visible object is.

4. Once you have determined what the closest fully visible object is, check if there are any small, distinct objects placed on top of or inside the fully visible object. These are referred to as secondary objects and should not include integral parts of the visible object itself. List these objects if applicable.

5. Once you've identified the objects, compile your list in the format:\\
\quad - \{VISIBLE\_OBJECT: [Object]\}\\
\quad - \{SECONDARY\_OBJECTS: [Object 1, Object 2, ..., Object N]\}

Examples:\\
\quad - For an empty scene: \{VISIBLE\_OBJECT: []\}, \{SECONDARY\_OBJECTS: []\}\\
\quad - For a scene with a table as the closest fully visible object and a book on top: \{VISIBLE\_OBJECT: [Table]\}, \{SECONDARY\_OBJECTS: [Book]\}

\# Guidelines

1. If there are multiple objects that seem to be equally distant from the camera and fully visible, pick one of them to be selected for \{VISIBLE\_OBJECT: []\}.

2. In your final list of objects, exclude objects that could be considered background elements, such as floors, walls, windows, doors, grass, rugs, or other structures. If the visible object is one of these background elements, pick the next closest fully visible object if there is one. If there isn't, state so.

3. When describing objects, try to keep the names of these objects between 1--3 words. Avoid unnecessary adjectives.

4. When compiling your lists, use the singular form of objects, even if there are multiple. Only include the name of the object in your compiled lists, NOT the explanations.

5. If there are no secondary objects on or inside the fully visible object, format the list as: \{VISIBLE\_OBJECT: [Object]\}, \{SECONDARY\_OBJECTS: []\}

6. If you are unsure if there are any objects, leave the lists empty: \{VISIBLE\_OBJECT: []\}, \{SECONDARY\_OBJECTS: []\}

\# Prompt

Using the provided instructions and image of a scene, identify the object closest to the camera that is fully visible and not occluded by any other objects. Then, determine if there are any secondary objects on or inside it. Compile your findings in the formats \{VISIBLE\_OBJECT: []\} and \{SECONDARY\_OBJECTS: []\}.
\end{promptbox}

\begin{promptbox}[title=Amodal Selection Output]

\vspace{0.5em}

\textbf{IMAGE:}

\begin{center}
\includegraphics[width=\linewidth]{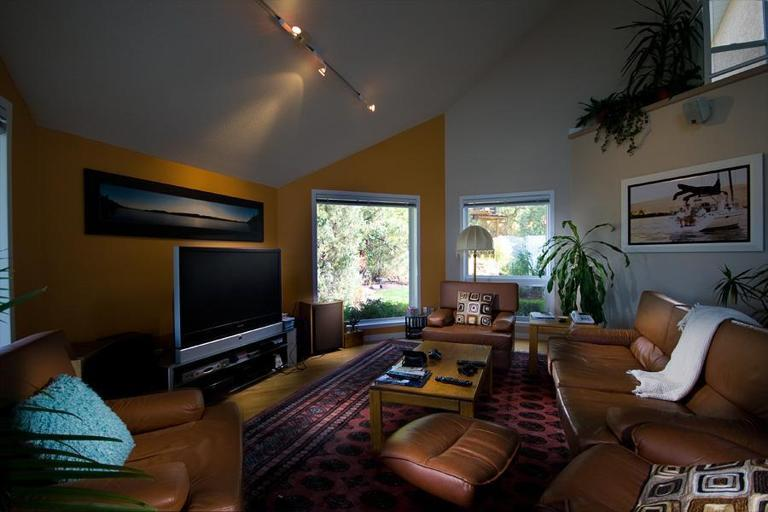}
\end{center}

\textbf{OUTPUT:}

Scene description:\\
A cozy living room with brown leather seating, a central wooden coffee table on a patterned rug, a large TV, windows with outdoor greenery, and several plants and wall art.

Closest fully visible object:\\
The small leather ottoman in the foreground, centered near the bottom.

Secondary objects on or inside it:\\
None visible.

\quad - \{VISIBLE\_OBJECT: [Ottoman]\}\\
\quad - \{SECONDARY\_OBJECTS: []\}

\end{promptbox}

\pagebreak
\subsection{Inpainting}

\begin{promptbox}[title=Inpainting Prompt]
\textbf{Prompt:} Remove the \{OBJ\_NAME\} inside of the red highlight.

\vspace{0.5em}

\centering
\includegraphics[width=\linewidth]{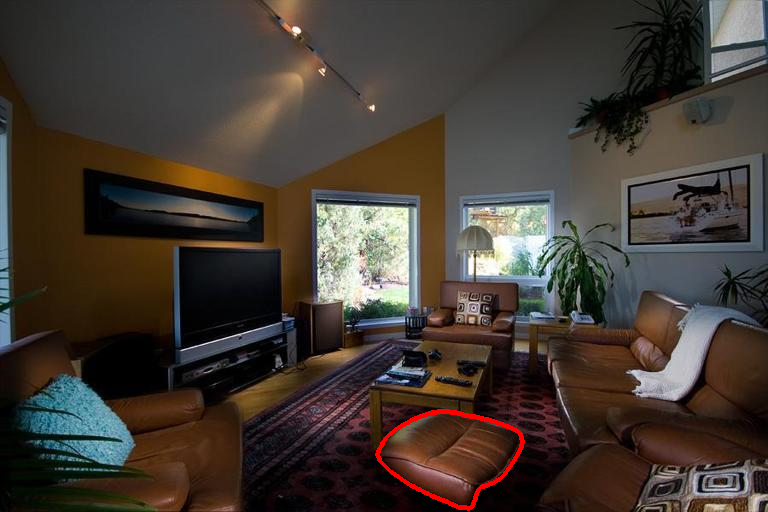}

\textbf{Output:}
\vspace{0.5em}

\centering
\includegraphics[width=\linewidth]{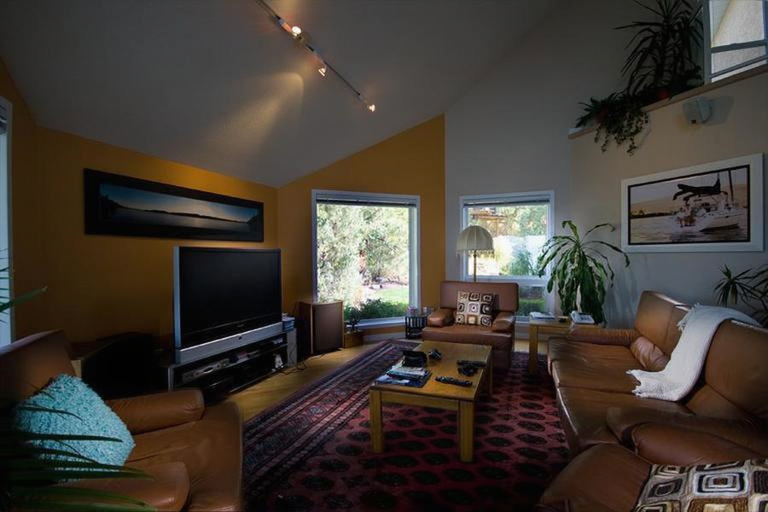}
\end{promptbox}

\section{Object Fitting Algorithms}

\begin{algorithm}
\caption{LEAST SQUARES}
\label{alg:sim3-least-squares}
\KwIn{$X_B \in \mathbb{R}^{N \times 3}$: points from rotated render (space B) \\
\hspace{1.7em}$X_A \in \mathbb{R}^{N \times 3}$: points from depth (space A)}
\KwOut{$T$: Sim(3) transform $[sR \;|\; t]$ mapping $B \to A$}

$\mu_B \gets \frac{1}{N} \sum X_B,\quad \mu_A \gets \frac{1}{N} \sum X_A$\;

$\tilde{X}_B \gets X_B - \mu_B,\quad \tilde{X}_A \gets X_A - \mu_A$\;

$\Sigma \gets \frac{1}{N}\tilde{X}_A^\top \tilde{X}_B$\;

$U,D,V^\top \gets \text{SVD}(\Sigma)$\;

$S_{fix} \gets \text{diag}(1,1,\text{sign}(\det(UV^\top)))$\;

$R \gets U S_{fix} V^\top$\;

$s \gets \frac{\text{trace}(S_{fix}D)}{\text{Var}(\tilde{X}_B)}$\;

$t \gets \mu_A - sR\mu_B$\;

Form $T \in \mathbb{R}^{4\times 4}$ with $T_{0:3,0:3} = sR$, $T_{0:3,3} = t$\;

\Return $T$\;
\end{algorithm}

\begin{algorithm}
\caption{ICP}
\label{alg:trimmed-icp}
\KwIn{$P_B$: source cloud from rotated render \\
\hspace{1.7em}$P_A$: target cloud from segmented depth \\
\hspace{1.7em}$v$: voxel size, $r$: NN radius, $\rho$: keep ratio, $T_{max}$: max iters}
\KwOut{$T$: Sim(3) transform $[sR \;|\; t]$ mapping $B \to A$}

$P_B \gets$ VoxelDownsample($P_B, v$)\;
$P_A \gets$ VoxelDownsample($P_A, v$)\;

Initialize scale $s$, rotation $R=I$, translation $t$ by centroid alignment\;

\For{$iter \gets 1$ \KwTo $T_{max}$}{
    $P_B^{now} \gets sR P_B + t$\;
    $Corr \gets$ NearestNeighbor($P_B^{now}, P_A, r$)\;

    \If{$|Corr| < N_{min}$}{\textbf{break}}

    Compute residuals $\{ \|p_B^{now} - p_A\| \}_{(p_B,p_A) \in Corr}$\;

    Keep top $K=\max(8, \lfloor \rho |Corr| \rfloor)$ pairs\;

    $X \gets$ source subset, $Y \gets$ target subset\;

    $(s,R,t) \gets$ \textbf{SIM3\_LEAST\_SQUARES}($X,Y$)\;

    \If{change in RMS $< \delta$}{\textbf{break}}
}
Form $T$ from $s,R,t$ and return\;
\end{algorithm}

\begin{figure*}
      \centering
      \includegraphics[width=\linewidth]{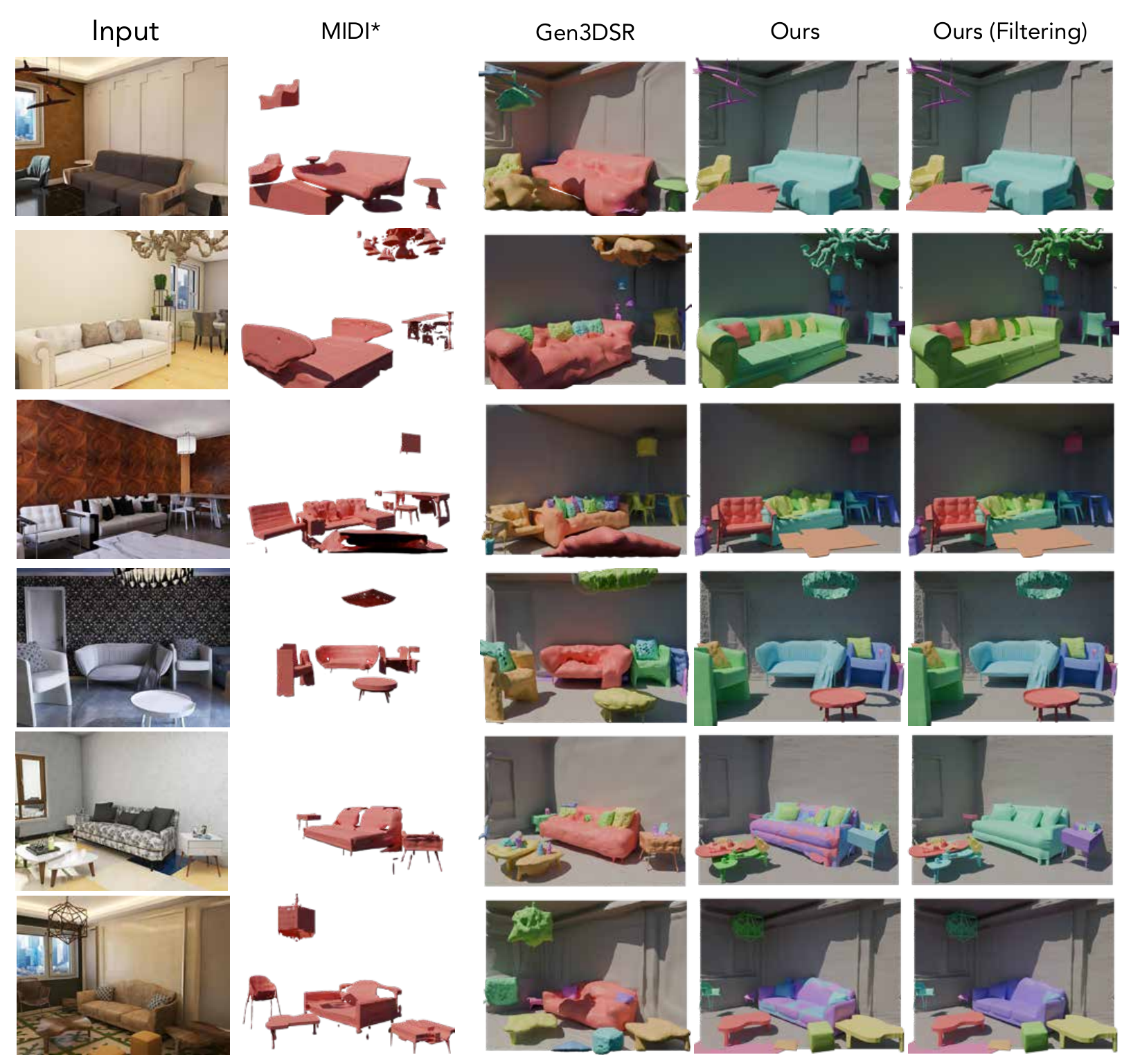}
      \caption{Qualitative comparison between MIDI*, Gen3DSR, ours, ours (w/ object filtering) on 3D front scenes.}
      \label{fig:front3d_gallery}
\end{figure*}

\end{document}